# Multimodal Trajectory Prediction for Autonomous Driving on Unstructured Roads using Deep Convolutional Network

Lei Li, Zhifa Chen, Jian Wang(✉), Bin Zhou, Guizhen Yu and Xiaoxuan Chen

Key Laboratory of Autonomous Transportation Technology for Special Vehicles, Ministry of Industry and Information Technology, School of Transportation Science and Engineering, Beihang University, Beijing 100191, China.
`{llsxyc, chenzhifa, wj1974, binzhou, yugz, 20374056}@buaa.edu.cn`

**Abstract.** Recently, the application of autonomous driving in open-pit mining has garnered increasing attention for achieving safe and efficient mineral transportation. Compared to urban structured roads, unstructured roads in mining sites have uneven boundaries and lack clearly defined lane markings. This leads to a lack of sufficient constraint information for predicting the trajectories of other human-driven vehicles, resulting in higher uncertainty in trajectory prediction problems. A method is proposed to predict multiple possible trajectories and their probabilities of the target vehicle. The surrounding environment and historical trajectories of the target vehicle are encoded as a rasterized image, which is used as input to our deep convolutional network to predict the target vehicle's multiple possible trajectories. The method underwent offline testing on a dataset specifically designed for autonomous driving scenarios in open-pit mining and was compared and evaluated against physics-based method. The open-source code and data are available at https://github.com/LLsxyc/mine_motion_prediction.git

## 1    Introduction

In recent years, with the rapid development of V2X, perception, decision and planning technologies in confined scenarios in open-pit mining sites, autonomous driving has also been effectively implemented to ensure efficient and safe completion of tasks for autonomous mining trucks [1]. However, there are still instances where autonomous trucks share roads and interact with human-driven vehicles. Therefore, accurate prediction of the trajectory for the surrounding human-driven vehicles is essential in autonomous driving system and will significantly benefit safety [2–4].

Within the unstructured road scenes of open-pit mining sites, as depicted in the bird's eye view shown in Figure 1, there exist notable differences when contrasted to structured urban roads. The unstructured road scenes have typical characteristics: the road has uneven boundaries and lacks clearly defined lane markings, although drivable and non-drivable parts are identifiable. Additionally, intersections lack of traffic devices such as traffic lights [5]. Vehicles in the open-pit mining sites are constrained solely by rules requiring them to drive on either the right or left side. When facing complex road conditions, the behavior (such as turning left, turning right, or going straight) of human-driven vehicles and their potential future motion become significantly more varied in comparison to driving on urban roads. Therefore, the task of predicting the target vehicle's multiple possible trajectories in unstructured road sceneries of the open-pit mining sites becomes difficult due to the constraints of limited vehicle states and map data.

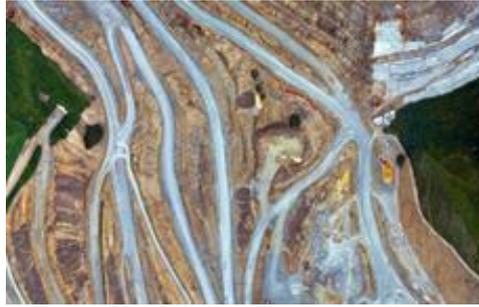

**Fig. 1.** Bird's Eye View (BEV) of unstructured road network in an open-pit mining.

Current research in the field of multimodal trajectory prediction of autonomous driving generally concentrates on structured road networks, such as metropolitan areas and highways. There are two widely used approaches: the physics-based methods [6, 7] and the learned-based methods [8–13]. Early studies predominantly employed physics-based method, which have shown effectiveness in short-term trajectory prediction but exhibit limitations in long-term predictions. Due to the rapid development of artificial intelligence, learning-based approaches have significantly improved in predicting long-term trajectories. Typically, these learning-based methods encode high-definition maps containing semantic information [14], such as lane markings, traffic signals, and pedestrian crossings, to generate trajectory predictions. However, predicting the target vehicle's multiple possible trajectories in unstructured road scenes in the open-pit mining sites presents a difficulty due to the limited vehicle statuses and map data available. Therefore, we propose a method that encodes maps, the surrounding environment of other vehicles, and vehicles' historical states into grid representations, which are then used as inputs for Convolutional Neural Networks (CNN). This enables the prediction of multiple possible trajectories of target vehicles and their probabilities.

The contributions are summarized as follows:

- A method is proposed for modeling scene context and encoding historical pose centered on the target vehicle for unstructured roads in the open-pit mining sites.
- A CNN-based method for trajectory prediction that accounts for the uncertainty of target vehicles, inferring multiple potential trajectories and their probabilities.
- Open-Source Implementation: An open-source code implementation of the proposed framework is provided, facilitating further research and development.

## 2 Related Work

In recent years, many methods have been proposed for predicting the future trajectories of other human-driven vehicles in the context of autonomous driving, with comprehensive overviews provided in many literature sources. We will first introduce commonly used engineering methods in real-world applications, followed by a discussion on deep learning approaches for trajectory prediction.

### 2.1 Physics-based methods

Accurately predicting the motion of participants is a key component of autonomous driving systems. In particular, predictions are closely linked to the planning and decision-making of autonomous vehicles, making the precise estimation of future conditions essential for their safe functioning. The majority of currently implemented autonomous driving systems employ established engineering techniques for trajectory prediction.

Common methods use vehicle dynamics or kinematics models, based on Constant Velocity, Constant Acceleration, Constant Turn Rate and Velocity, and Constant Turn Rate and Acceleration models. They employ Kalman Filter (KF) techniques to predict the future state of vehicles, including position, heading angle, speed, and acceleration. The KF method can handle process noise by using a predict-and-update approach to calculate the state and covariance matrix at each time step [2].

Vasileios Lefkopoulos et al. [7] implemented a longitudinal intention prediction based on an interaction model, which takes into account a surrounding environment that is free from collisions, while this approach has certain computational benefits, its evaluation outcomes are less favorable when compared to learning-based methods. Atsushi Kawasaki [6] constructed a cubic function speed model based on the geometry of intersections and applied it to the Extended Kalman Filter (EKF) for predicting future trajectories. However, this approach presupposes that the turning intentions of vehicles, involving both left and right turns, are pre-established. The method only predicts the speed of vehicles at the speed control points entering the intersection, resulting in a single deterministic trajectory prediction. Thus, this model struggles to provide more comprehensive information to downstream processes.

### 2.2 Learned-based Methods

Physics-based techniques frequently encounter difficulties in addressing the intricacies of traffic in real-world scenarios. As a result, a growing number of researchers are considering the implementation of learned-based models as a solution.

Classic machine learning methods suitable for trajectory prediction in autonomous vehicles include Gaussian Processes (GP), Support Vector Machines, Hidden Markov Models (HMM), and Dynamic Bayesian Networks. Guo et al.[8] formulated a model for multi-vehicle interaction scenarios using GP, generating a mixture from naturalistic data via non-parametric Bayesian learning. Holger Berndt et al.[9] used the steering angle and global coordinates as inputs for HMM to predict driver maneuvers. However,

HMM-based methods struggle to account for the impact of interactions in real traffic scenarios.

Deep learning has rapidly advanced in the autonomous driving industry. Zyner et al. used Recurrent Neural Networks with a weighted Gaussian Mixture Model for prediction, obtaining parameters through a three-layer LSTM encoder-decoder to extract the most probable set of trajectories and the final result was obtained through clustering [10]. Cui et al. used grid images as inputs combined with motion models to complete vehicle trajectory prediction [11]. Kawasaki et al. considered lane interaction information and input it into an LSTM-based encoder-decoder framework, integrating a KF-based motion model method [12]. Additionally, Nikhil et al. argued that since trajectories exhibit strong spatiotemporal continuity, using CNNs for trajectory prediction is better than RNNs [13]. They employed a sequence-to-sequence structure, using historical trajectories as input, maintaining temporal continuity through convolutional layers stacked after fully connected layers, and outputting future trajectories via fully connected layers. Therefore, choosing CNNs for trajectory prediction is justified.

## 3 Proposed Approach

This chapter will discuss the proposed approach for the target vehicle's multimodal trajectory prediction in open-pit mining sites. The discussion is divided into four sections: problem settings, model inputs, multimodal trajectories generator, and loss function.

### 3.1 Problem Settings

Data can be obtained through V2X technology, LiDAR, and other sensor devices. These devices are used by detection and tracking systems in autonomous driving systems to provide estimated states $s = [x, y, \theta, v, a, \omega]$ of surrounding other vehicles, where $(x, y, \theta)$ represent the position and orientation in Cartesian coordinates, $v$ denotes the vehicle's instantaneous velocity, $a$ represents the vehicle's acceleration, and $\omega$ is the vehicle's yaw rate. The system can output the participants' historical states $S_T = \{s_{T-k+1}, s_{T-k+2}, s_{T-k+3}, \ldots, s_T\}$ with a fixed time interval between the consecutive trajectory (for example, output at a frequency of 2Hz, which corresponds to a time interval of 0.5 seconds, $k$ representing the length of the historical sequence, set to 6 in this article, $s_i$ represents the output of the state of target vehicle at time step $i$). The task of trajectory prediction is to obtain a sequence of future states $[s_{T+1}, s_{T+2}, \ldots, s_{T+H}]$ using the historical trajectories and pre-obtained high-definition map information, where $H$ represents the continuous time step of predicted future states. Additionally, we simplify the task to infer the future position of the target vehicle, denoted as $[\boldsymbol{x}_{T+1}, \boldsymbol{x}_{T+2}, \ldots, \boldsymbol{x}_{T+H}]$, where $\boldsymbol{x}_i$ represents the vehicle's position $[x, y]$.

### 3.2 Model Inputs

Model inputs are processed as follows: firstly, only the interested vehicle is rasterized, taking into account the limited interaction scenes in mining sites and the state is used to model the dynamic context at each time step. Map data containing drivable and non-drivable areas are used to represent the static context, with these areas being represented as polygons. Target vehicle at the time step $T$ is rasterized into an RGB image. This rasterized image along with the current state of the other vehicles is used as inputs for CNN to infer trajectory outputs. The detailed network architecture will be illustrated in Section 3.3.

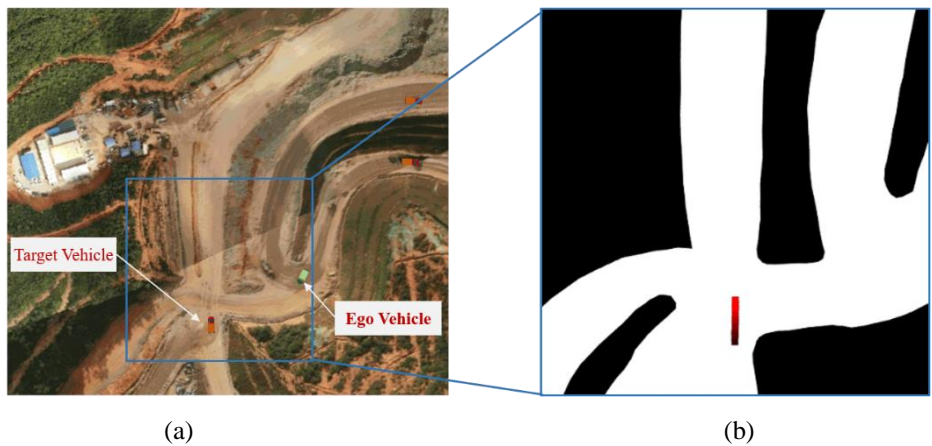

<div align="center">(a)               (b)</div>

**Fig. 2.** Rasterized image example. (a) the actual instance in the mining site. (b) depicts non-drivable regions as black patches, drivable regions as white patches, and the agent's historical trajectories are indicated by faded red color.

Considering the image size and the ability to represent details accurately, the pixel resolution is set to 0.1m (10 pixels represent 1m). To represent the context of the agent at the time step $T$, an $n \times n$ image $I_T$ is created with the agent positioned at pixel $(w,h)$, where $w$ and $h$ represent the width and the height measured from the bottom left corner of the image. Given the relatively large size of vehicles in the mining sites, $n$ is set to 1200. The agent is positioned at $(w,h)=(600,300)$ and is depicted in a coordinate system, with its color displayed as red. To capture the historical features of the target vehicle's past positions, the bounding boxes of the vehicle's historical trajectory sequence are converted into a rasterized format and displayed on the top layer of the map. Each historical polygon has the same color as the agent, but its brightness levels gradually decrease to produce a fading effect. The brightness level of the time step $T-K$ is set to $\max(0, 1-K\delta)$, where $K=0,1,\ldots,k-1$ and $\delta=0.1$. The result is shown in the Figure 2.

### 3.3 Multimodal Trajectories Generator

The CNN-based model is used for predicting $M$ possible future trajectory sequences $\left\{ \left[ \boldsymbol{x}_{m(T+1)}, \boldsymbol{x}_{m(T+2)}, \ldots, \boldsymbol{x}_{m(T+H)} \right] \right\}_{m=1,2,\ldots,M}$ along with corresponding probabilities $p_m$ for each sequence, where $\sum_m p_m = 1$ and $m$ represents the index of the trajectory sequence. An agent-centric RGB rasterized image with a resolution 0f 0.1m, containing map semantic information and historical trajectory, serve as the input to CNN for extracting multi-scale features. These features, concatenated with the vehicle's motion state $(v, a, \boldsymbol{\omega})$ are used as inputs to MLP to obtain multimodal trajectory. Softmax is the applied to obtain the top M trajectories probabilities. Finaly, the output consists of M modes for future trajectories, each with its corresponding probability. In total, there are $(2H+1)M$ outputs. During this process, the CNN model can adopt ResNet [15] or MobileNet-v2 [16] as backbone. In our experiments, MobileNet-v2 is utilized as the backbone.

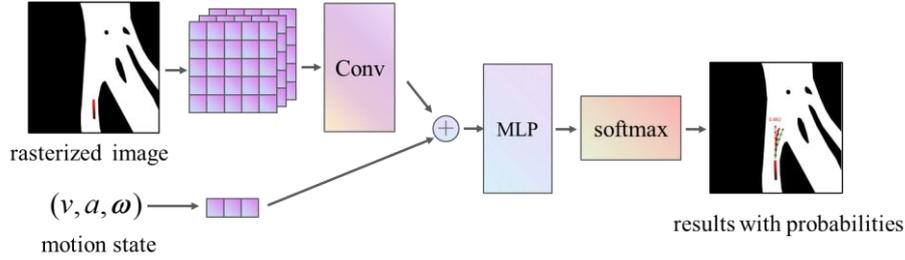

**Fig. 3.** Network architecture

### 3.4 Loss Functions

In the section, the design of the loss function for modeling multi-modal trajectory prediction problems are discussed. The single-mode loss for time step $T$ and mode $m$ is defined as the average displacement error between the ground truth trajectory $\boldsymbol{x}_{mT}$ and the prediction trajectory $\boldsymbol{x}_{mT}$ (i.e. $\ell_2 - norm$),

$$L\left( \boldsymbol{x}_T, \boldsymbol{x}_{mT} \right) = \frac{1}{H} \sum_{h=1}^{H} \left\| \boldsymbol{x}_T^h - \boldsymbol{x}_{mT}^h \right\|_2 \tag{1}$$

where $\boldsymbol{x}_T^h$ and $\boldsymbol{x}_{mT}^h$ represent the x-position and y-position of $\boldsymbol{x}_{nT}$ and $\boldsymbol{x}_{mT}$ at time $h$.

The common loss function design for multi-modal trajectory prediction is the Mixture-of-Experts (ME) loss, which is given by $\mathcal{L}_T^{ME} = \sum_{m=1}^{M} p_m L\left( \boldsymbol{x}_T, \boldsymbol{x}_{mT} \right)$. As indicated in [11], the ME loss often leads to mode collapse issues, therefore, a trajectory distance function $dist\left(\boldsymbol{x}_T, \boldsymbol{x}_{mT}\right)$ that considers the distance and angle differences between prediction and the ground truth to determine the optimal mode $m^*$,

$$m^* = \underset{m \in \{1,\dots,M\}}{\arg\min} \, dist(\boldsymbol{x}_T, \boldsymbol{x}_{mT}) \tag{2}$$

Once the optimal mode $m^*$ is selected, the final loss function can be defined as follows:

$$\mathcal{L}_T = \mathcal{L}_T^{class} + \alpha \sum_{m=1}^{M} I_{m=m^*} L\left(\boldsymbol{x}_T, \boldsymbol{x}_{mT}\right) \tag{3}$$

where $I_c$ is the binary indicator function, i.e. $I_c = \begin{cases} 1, & \text{c is true} \\ 0, & \text{c is false} \end{cases}$, $\mathcal{L}_T^{class}$ is defined as the classification cross-entropy loss, given by

$$\mathcal{L}_T^{class} = -\sum_{m=1}^{M} I_{m=m^*} \log p_m \tag{4}$$

and $\alpha$ is a hyperparameter used to balance these two losses. During the loss calculation process, we only focus on the probability and loss of the optimal mode, and update them during training.

Finally, the CNN parameters $\Theta$ are trained with respect to (3) to minimize the loss on the training data.

$$\Theta^* = \underset{\boldsymbol{\theta}}{\arg\min} \sum_{t=1}^{T} \mathcal{L}_t \tag{5}$$

## 4 Experiments

### 4.1 Datasets and Experiment Setup

We collected real data from autonomous and manually driven pickups and mining trucks at two different open-pit mining sites. The raw vehicle trajectory data was collected based on positioning data and recorded at a rate of 10Hz, capturing the vehicle's motion state, including latitude and longitude position, heading angle, speed, and acceleration, describing the vehicle's physical state during travel. Additionally, we collected aerial images of the mining areas using drones. Based on these aerial images and actual maps of the mining areas, we divided the maps into drivable and non-drivable regions using polygon layers.

Different origins were selected based on the mining areas, and the semantic information of the open-pit mining sites maps was described in polygon format. The vehicle trajectories were calculated relative to the chosen origins through coordinate system transformations, and the real data was segmented to generate multiple instances.

Considering the practical needs of trajectory prediction, we downsampled the historical state estimation to 2Hz, resulting in a total of 16,104 frames of pre-training data. For trajectory prediction, we aim to predict the next 6 seconds ($H = 6$). The dataset was divided into training, validation, and test sets with a ratio of $7 : 1.5 : 1.5$.

Based on these, we chose to implement our model using PyTorch and deployed it on a GPU. The hyperparameters are set with a batch size of $64$ and an initial learning rate of $10^{-4}$. Additionally, we compare the performance of the following methods on the dataset:

- Estimating future states using EKF;
- Single-modal trajectory prediction method [17].

### 4.2 Qualitative Results

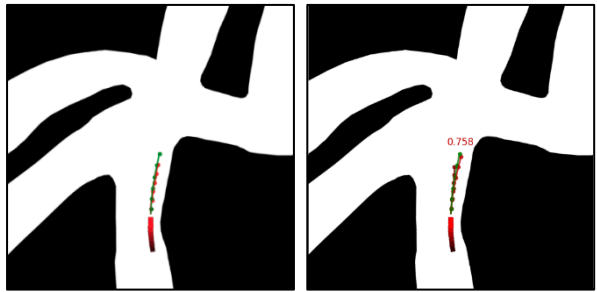

**Fig. 4.** Trajectory prediction results of the EKF (on the left) and Ours with modes $m = 2$ (on the right). The green line refers the ground truth, while the red lines refer the prediction trajectories and as the brightness of the red color increases, the probabilities drop. The number indicates the highest probability value.

Figure 4 and Figure 5 show visualizations of different methods used on the same scenario. Figure 4 demonstrates that EKF is unable to identify changes in the scene and merely relies on the current state for prediction, while our method accurately captures the right turn maneuver. Furthermore, as shown in Figure 4, when the vehicle accelerates through an intersection, our method can accurately predict the speed changes of vehicles accelerating through intersections, indicating that the acceleration patterns at intersections remain constant.

Figure 5 illustrates that with the number of modes increases, more possible trajectories of the vehicle going through the intersection are shown. When the modes are set to 5, the closest optimal result to the ground truth is achieved. This indicates that it is necessary to anticipate more modes in order to capture a broader range of conceivable trajectories.

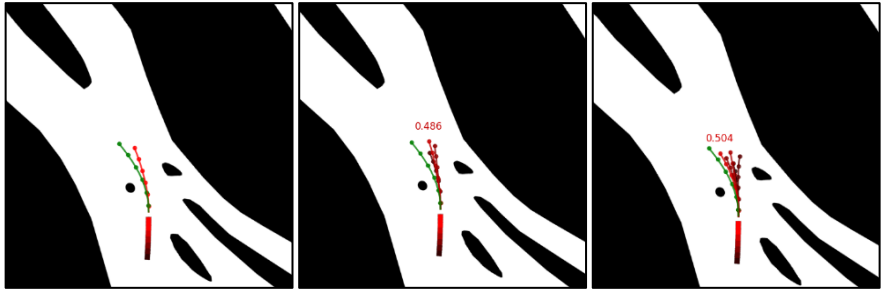

**Fig. 5.** The trajectory prediction results of EKF, Ours with modes $m=3$ and Ours with modes $m=5$ are displayed from left to right.

### 4.3 Quantitative Results

**Table 1.** The evaluation results from different methods with the best results noted in **bold**.

| Method | Metrics | | |
|---|---|---|---|
| | minADE | minFDE | missRate |
| EKF | 2.195 | 5.485 | \ |
| Single-modal [17] | 1.287 | 3.282 | 0.614 |
| Ours with $m=2$ | 1.041 | 2.604 | 0.493 |
| Ours with $m=3$ | 0.904 | 2.228 | 0.373 |
| Ours with $m=5$ | **0.778** | **1.865** | **0.281** |

Table 1 shows that the ideal results for minADE, minFDE, and missRate are obtained when modes $m=5$, suggesting a significant improvement of our method on complex road conditions in the open-pit mining sites. It is noted that the single-modal trajectory prediction method based on [17] shows 1.7 times increase in minADE and minFDE on the mining site dataset. This indicates that the performance of utilizing a deep convolutional network for trajectory prediction is outstanding with respect to average displacement error and final distance error. This finding suggests that convolutional network-based trajectory prediction methods effectively recognize the surrounding environment and generate results that better match real-world scenarios.

Ours well captures the intricate multi-modal properties of intersections, as evidenced by the consistent decline in the minADE and minFDE metrics as the number of modes grows. Additionally, the missRate is the best and the prediction error is lower than other approaches, demonstrating that the prediction hit rate increases steadily even as the number of modes increases. This proves that in the open-pit mining sites, the method obtains the best performance in trajectory prediction.

### 4.4 Improvement

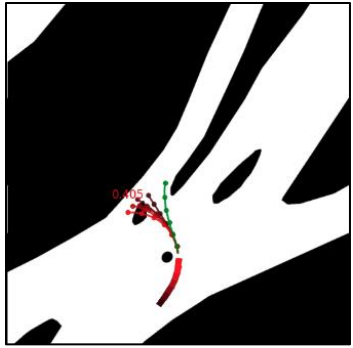

**Fig. 6.** Prediction trajectories fall into non-drivable areas with $m=5$

Trajectory prediction problems are made more challenging by the complicated irregular intersections and unstructured roads that are an inherent characteristic of mining sites. As shown in Figure 6, in such cases, almost half of the results pass through locations that are non-drivable areas. The probabilities of these trajectory results should be zero since they are unacceptable. To prevent undesirable outcomes, the next improvement should involve deleting these results or adding trajectory priors.

## 5 Conclusion

Due to the complex unstructured roads in the mining sites, autonomous vehicles must take into account the inherent diversity of future trajectories of surrounding human-driven vehicles to ensure safe driving. A method to generate multimodal results for trajectory prediction is proposed in this paper. A rasterized image containing historical trajectories and the surrounding environment is generated, and a CNN-based model is used to output possible trajectories and their probabilities. EKF and single-modal trajectory prediction methods were discussed and compared with our method. Results indicate that the proposed method performs better on unstructured roads in the open-pit mining sites.

**Acknowledgements** This work is partially supported by the National Key Research and Development Program of China (2022YFB4703700).